\documentclass[sigplan,screen]{acmart}
\AtBeginDocument{%
  }

\setcopyright{acmlicensed}
\copyrightyear{2018}
\acmYear{2018}
\acmDOI{XXXXXXX.XXXXXXX}
\acmConference[Conference acronym 'XX]{Make sure to enter the correct
  conference title from your rights confirmation email}{June 03--05,
  2018}{Woodstock, NY}
\acmISBN{978-1-4503-XXXX-X/2018/06}

\copyrightyear{2025}
\acmYear{2025}
\setcopyright{rightsretained}
\acmConference[GAIB 2025]{2025 International Conference on Generative Artificial Intelligence for Business}{August 04--06, 2025}{Hongkong, China}
\acmBooktitle{2025 International Conference on Generative Artificial Intelligence for Business (GAIB 2025), August 04--06, 2025, Hongkong, China}
\acmPrice{}
\acmDOI{10.1145/3766918.3766953}
\acmISBN{979-8-4007-1602-7/25/08}

\begin{document}

\title{Incremental Hybrid Ensemble with Graph Attention and Frequency-Domain Features for Stable Long-Term Credit Risk Modeling}

\author{Jiajing Wang}
\email{wangjj1502@126.com}
\affiliation{%
  \institution{Columbia University in the City of New York}
  \city{New York}
  \country{USA}
}

\renewcommand{\shortauthors}{Trovato et al.}

\begin{abstract}
  Predicting long-term loan defaults is hard because borrower behavior often changes and data distributions shift over time. This paper presents HYDRA-EI, a hybrid ensemble incremental learning framework. It uses several stages of feature processing and combines multiple models. The framework builds relational, cross, and frequency-based features. It uses graph attention, automatic cross-feature creation, and transformations from the frequency domain. HYDRA-EI updates weekly using new data and adjusts the model weights with a simple performance-based method. It works without frequent manual changes or fixed retraining. HYDRA-EI improves model stability and generalization, which makes it useful for long-term credit risk tasks.
\end{abstract}

\ccsdesc[500]{Computing methodologies~Supervised learning by classification}
\ccsdesc[500]{Computing methodologies~Ensemble methods}
\ccsdesc[500]{Computing methodologies~Feature selection}
\ccsdesc[500]{Computing methodologies~Online learning settings}
\keywords{Loan default prediction, concept drift, ensemble learning, incremental training, feature engineering}

\maketitle

\section{Introduction}
Predicting long-term loan defaults is important in credit risk control. But this task is difficult. Borrowers often change their repayment behavior over time. Also, economic environments shift, and financial products evolve. These changes cause concept drift, where the patterns in the data change slowly or suddenly. Many traditional credit scoring models are not designed for this. They are trained once and then used for a long time. So they often become less accurate as time goes on.

To solve this, a model must do more than just learn once. It should update regularly. It should also learn new data while remembering useful past knowledge. In addition, it should use features that show not just individual values but also how borrowers are connected, how their actions interact, and how their behavior changes over time. This means it needs a way to use graph structures, cross features, and time-frequency signals.

In this paper, we propose HYDRA-EI. It is a hybrid framework that uses multiple models together. It builds three types of features. First, it uses graph attention to find links between borrowers. Then, it uses automatic methods to create cross features. These help the model find complex patterns. Finally, it uses tools like Fourier and wavelet transforms to find periodic trends and short-term changes in the data.

The model includes three learners: LightGBM, CatBoost, and DenseLight\textsuperscript{+}. Each model is updated weekly with new data. Their outputs are combined using a Bayesian gating method. This method gives more weight to models that work better on recent data. The whole system updates smoothly without starting over. This helps the model stay accurate when the data keeps changing.

HYDRA-EI gives a full method for credit default prediction. It adapts to drift, keeps learning, and uses strong features. This makes it useful for real financial systems that need long-term and stable risk prediction.

\section{Related Work}

Concept drift is a common problem. Souza et al.\cite{souza2020challenges} pointed out that stream models often fail when the data changes. Jiang et al. \cite{jiang2024robustkv} propose RobustKV, which enhances model robustness by evicting low-importance tokens from KV caches to mitigate adaptive jailbreak attacks, offering a complementary perspective on maintaining stability under adversarial distribution shifts.

GNNs are useful for time and structure modeling. Jin et al.\cite{jin2024survey} reviewed GNNs for time data. Wang et al.\cite{wang2024latent} propose an LVMTL that models estimated health indices as latent variables to capture dependency and heterogeneity and employ QOIEM (QP + MAP-EM) for robust parameter estimation under missing data. Luo \cite{luo2025fine} presents TriMedTune, a triple-branch framework (HVPI, DATA, MKD-UR) with LoRA fine-tuning, dynamic prompt sampling, and mixed-precision optimization to boost multimodal brain CT diagnosis and report generation.

Some works use models that update. Jain et al.\cite{jain2025incremental} designed an update-friendly model. Shyaa et al.\cite{shyaa2023enhanced} added drift detection using simple genetic methods.Zhang and Hart\cite{zhang2023effect} show inverse-gamma prior shape controls posterior concentration; adopting small-shape (weak) priors stabilizes HYDRA-EI’s Bayesian gating under noisy or scarce validation data.

Guo and Yu \cite{guo2025privacypreservenet} introduce PrivacyPreserveNet, a multilevel privacy-preserving framework for multimodal LLMs that combines differential privacy-enhanced pretraining, privacy-aware gradient clipping, and noise-injected attention to protect sensitive text, image, and audio data without degrading performance.Chen \cite{chen2024coarse} proposes a SLAM-based coarse-to-fine reconstruction with a Transformer multi-view matcher that improves feature matching and reprojection error; integrating its transformer matching into HYDRA-EI’s Graph Feature Synthesizer would improve node alignment and relational-embedding robustness.

These papers each solve part of the problem. But few models combine structure, time, and drift-handling. HYDRA-EI puts these parts together. It builds strong features, learns with each update, and balances models over time.

\section{Methodology}

\begin{figure}[htbp]
    \centering
    \includegraphics[width=0.5\textwidth]{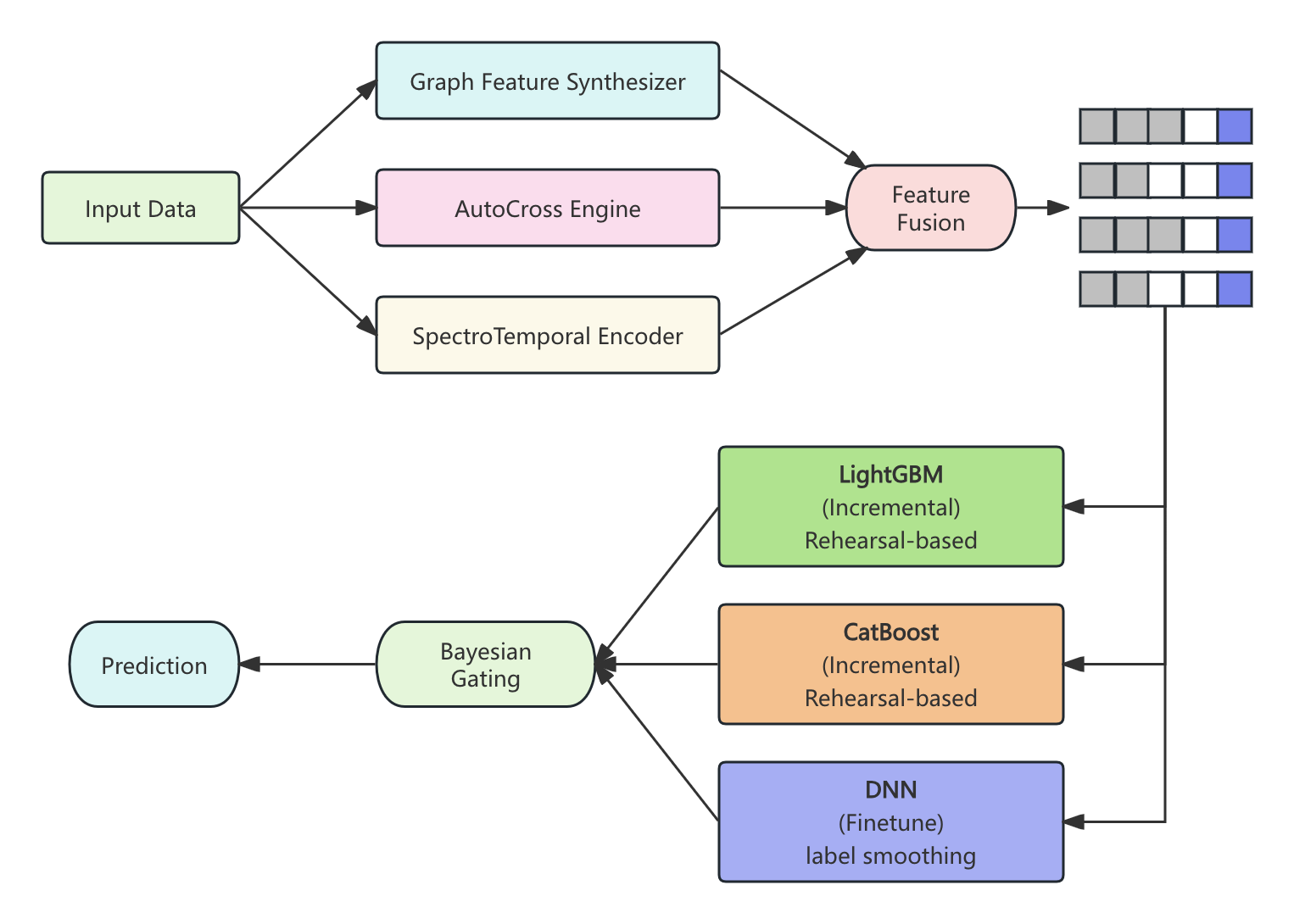}
    \caption{Overview of the HYDRA‐EI framework. The architecture consists of three models.}
    \label{fig:HYDRA‐EI}
\end{figure}

We present \textbf{HYDRA‐EI}, framework with ensemble incremental learning—for robust long-horizon loan-default prediction. HYDRA-EI integrates three core models—LightGBM, CatBoost, and a DNN within a novel performance-adaptive ensemble structure. The system features a Graph Feature Synthesizer to extract topological relations, an AutoCross engine for high-order feature evolution, and a SpectroTemporal Encoder that combines frequency and time-domain behavioral cues. Incremental training enables the tree models to handle concept drift via rehearsal-based updating, while the DNN is fine-tuned with label smoothing and gated SE blocks. A performance-aware Bayesian gating mechanism adaptively combines model predictions based on rolling validation scores. Extensive experiments confirm HYDRA-EI’s superior Gini stability and generalization, offering a strong and adaptive solution to dynamic credit risk modeling. The pipeline in Figure \ref{fig:HYDRA‐EI}

\subsection{Multistage Feature Engineering Module}

Feature engineering is one of the most influential aspects of model performance in tabular learning problems. We propose a three-stage feature enhancement strategy that captures relational, nonlinear, and periodic aspects of client behavior.

\subsection{Graph Feature Synthesizer}

In real-world finance data, clients are not independent entities—they share employers, merchants, and sometimes locations. We leverage this implicit connectivity through a Graph Attention Network (GAT)-based module. Specifically, clients are treated as nodes, and edges are formed if they share categorical identifiers, such as `WORK\_TYPE` or `MERCHANT\_GROUP`.

Each node aggregates contextual information from its neighbors through multi-head attention mechanisms:
\begin{align}
\mathbf{h}_i^{(1)} &= \sigma\left(\sum_{j \in \mathcal{N}(i)} \alpha_{ij}^{(0)} \mathbf{W}^{(0)} \mathbf{x}_j\right),\\
\mathbf{h}_i^{(2)} &= \sigma\left(\sum_{j \in \mathcal{N}(i)} \alpha_{ij}^{(1)} \mathbf{W}^{(1)} \mathbf{h}_j^{(1)}\right),
\end{align}
where $\alpha_{ij}^{(\ell)}$ is a softmax-normalized attention coefficient. A trick we found useful here is to restrict neighborhood size by hashing high-cardinality fields to avoid memory explosion and over-smoothing.

\begin{figure}[htbp]
    \centering
    \includegraphics[width=0.5\textwidth]{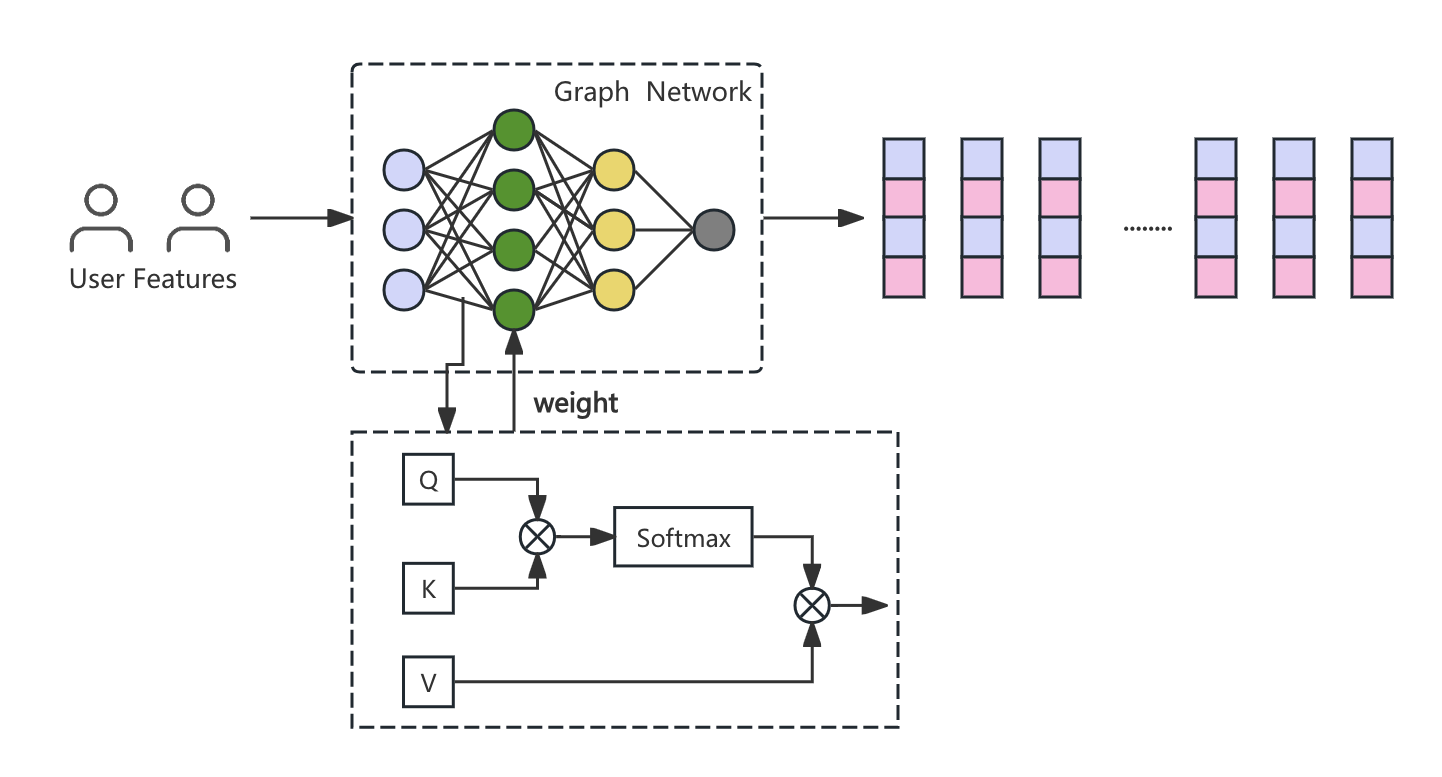}
    \caption{The architecture consists of Graph Feature Synthesizer.}
    \label{fig:graph}
\end{figure}

The final node embedding $\mathbf{z}_i$ is concatenated with other features for downstream use. Graph-based representation improves prediction for clients with limited historical features by borrowing contextual signals from similar profiles. The pipeline of Graph Feature Network is show in Figure \ref{fig:graph}

\subsection{AutoCross Feature Evolution}

Classical feature crosses (like \texttt{income \(\times\) spending}) are effective but labor-intensive. We automate this using a dynamic evolutionary engine, which proposes nonlinear feature interactions $x_{uv}^{(k)} = f_k(x_u, x_v)$ and evaluates them based on holdout loss drop:
\begin{equation}
\Delta \mathcal{L}_{val} = \mathcal{L}_{base} - \mathcal{L}_{base + feat}.
\end{equation}

A key challenge we faced was overfitting from sparse high-order interactions. To mitigate this, we regularized interaction proposals with occurrence thresholds (minimum frequency) and relied on LightGBM's feature importance to prune redundant crosses.

\subsection{SpectroTemporal Encoder}

Client behavior often exhibits periodicity, such as salary inflow or loan repayment. To capture this, we applied both discrete Fourier transform (DFT) and discrete wavelet transform (DWT) to weekly aggregated features. These encode both global frequency content and local anomalies:
\begin{align}
F_{i,\omega} &= \frac{1}{T} \left| \sum_{t=1}^{T} x_{i,t} e^{-j 2 \pi \omega t / T} \right|,\\
W_{i,\ell} &= \sum_{t=1}^{T} x_{i,t} \psi_{\ell}(t).
\end{align}

\begin{figure}[htbp]
    \centering
    \includegraphics[width=0.5\textwidth]{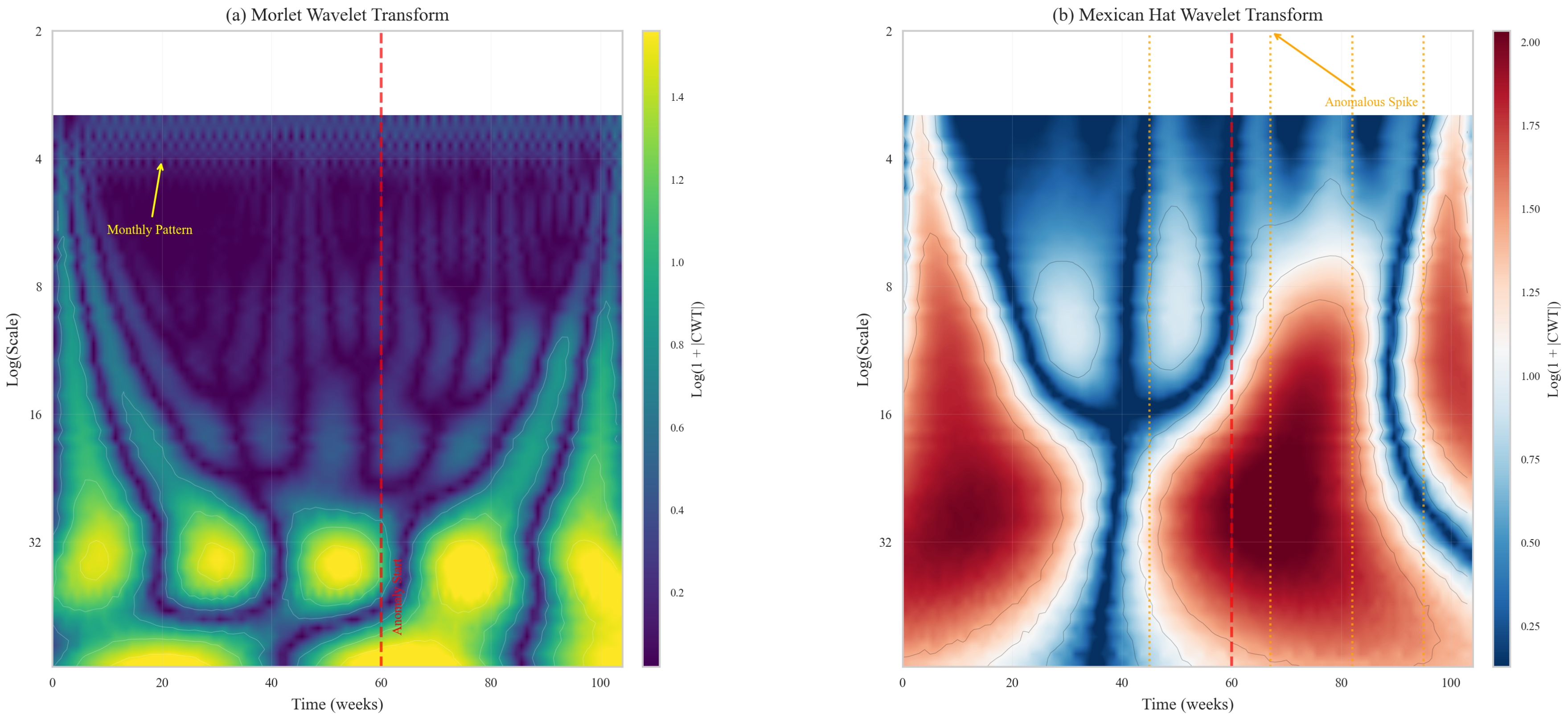}
    \caption{Wavelet transform analysis of client financial behavior over 104 weeks.}
    \label{fig:Encoder}
\end{figure}

Wavelet families were benchmarked; Morlet and Mexican Hat performed best. These transforms improved classification of users with regular but subtle risk behavior (e.g., slowly increasing expense trends). We filtered frequency bins using variance filtering to reduce dimensionality.The time series (top) shows normal (green) and risk (red) periods with anomalous spikes show in Figure \ref{fig:Encoder}

\subsection{Base Models}

We adopted a heterogeneous ensemble of three powerful base models: LightGBM, CatBoost, and DenseLight\textsuperscript{+}. Each model captures complementary data patterns.

\subsubsection{LightGBM with GOSS}

LightGBM is known for its speed and handling of large feature sets. We employed GOSS (Gradient-based One-Side Sampling) to focus training on high-gradient samples, improving learning efficiency. A critical trick was to fine-tune the \texttt{min\_data\_in\_leaf} and \texttt{max\_bin} parameters to balance underfitting on early folds and overfitting on drifted features.

The split criterion is:
\begin{equation}
\Delta H = \frac{G_L^2}{H_L + \lambda} + \frac{G_R^2}{H_R + \lambda} - \frac{(G_L + G_R)^2}{H_L + H_R + \lambda}.
\end{equation}

Categorical features were converted using ordered target encoding with noise injection to avoid leakage.

\subsubsection{CatBoost with Incremental Ordered Boosting}

CatBoost naturally handles categorical features and ordered boosting. However, long-horizon data causes concept drift. We implement an incremental learning scheme using CatBoost’s native support for warm-start:
\begin{equation}
\theta^{(t)} \leftarrow \theta^{(t-1)} - \eta \frac{\sum_{i \in \mathcal{B}_t} g_i^{(t)}}{\sum_{i \in \mathcal{B}_t} h_i^{(t)} + \lambda}.
\end{equation}

We found that refreshing the training set with a rehearsal buffer (random samples from past weeks) helps stabilize loss under distributional shifts. To further prevent leakage, we disabled `one-hot max size` and relied purely on permutation-driven encoding.

\subsubsection{DenseLight\textsuperscript{+} Neural Network}

Our DNN uses gated squeeze-and-excitation (SE) layers to emphasize informative features:
\begin{align}
\mathbf{s}^{(\ell)} &= \sigma\left(W_s^{(\ell)} \text{GAP}(\mathbf{h}^{(\ell-1)})\right),\\
\mathbf{h}^{(\ell)} &= \mathbf{h}^{(\ell-1)} + \mathbf{s}^{(\ell)} \odot \phi\left(W_f^{(\ell)} \mathbf{h}^{(\ell-1)}\right),
\end{align}
where $\phi$ is GELU. This structure improved both convergence and resilience to sparse features. A key challenge was balancing feature scale; we introduced layer-wise normalization to ensure numerical stability across batches.

To combat label imbalance and prediction confidence overfitting, we employed label smoothing:
\begin{equation}
\mathcal{L}_{DL+} = -\frac{1}{N} \sum_{i=1}^{N} \left[ \bar{y}_i \log p_i + (1 - \bar{y}_i) \log (1 - p_i) \right],
\end{equation}
with $\bar{y}_i = (1 - \varepsilon) y_i + \varepsilon / 2$ and $\varepsilon = 0.1$.

\subsection{Incremental Training Strategy}

Training was organized by weekly epochs, from earliest to latest (\(\mathcal{T}_1 \prec \cdots \prec \mathcal{T}_K\)). In each round:
\begin{enumerate}
    \item Construct new training data $\mathcal{B}_k$ from week $\mathcal{T}_k$ and mix with replay buffer $\mathcal{R}$.
    \item Update CatBoost and LightGBM incrementally via warm start.
    \item Fine-tune DenseLight\textsuperscript{+} on $\mathcal{B}_k$ with early stopping.
    \item Recalculate performance-aware ensemble weights (see next subsection).
\end{enumerate}

One issue we encountered was catastrophic forgetting in tree models. Using a 1:1 ratio of new to old samples in $\mathcal{R}$ was found to yield the best generalization stability.

\subsection{Bayesian Performance-aware Ensemble Gating}

Traditional ensemble methods use static averaging. However, we found that the performance of each base model varies significantly across time. To solve this, we adaptively weight model outputs using a Bayesian gate:
\begin{equation}
\alpha_m(t) = \frac{\beta_m + \exp(-\ell_m(t))}{\sum_{j} \beta_j + \exp(-\ell_j(t))},
\end{equation}
where $\ell_m(t)$ is the loss on validation fold at epoch $t$, and $\beta_m$ is a prior score.

This gate allows us to boost underperforming learners less when data drifts or categorical sparsity changes. We recalibrate weights every epoch, and apply them to generate the final prediction:
\begin{equation}
p_i^* = \sum_{m \in \{CB, LGB, DL+\}} \alpha_m(t_i) \cdot p_i^{(m)}.
\end{equation}

\subsection{Global Optimization Objective}

Our total training objective balances each model’s cross-entropy loss and a consistency regularizer:
\begin{equation}
\mathcal{J}_{total} = \sum_{m} \mathcal{L}_m + \gamma \cdot \text{KL}(p^* \| p^{target}),
\end{equation}
where $p^{target}$ is the soft output from the best expert under a temperature-scaled softmax. This encourages smoother decision boundaries and harmonization among base learners.

\section{Feature Engineering}

Combining statistical, temporal, interaction, and relational feature transformations in a modular pipeline enhances generalization, robustness, and client behavior representation.

\subsection{Statistical and Volatility Features}

We computed summary statistics such as mean, variance, maximum, and trend over fixed-length time windows:
\begin{align}
\text{Mean}(x) &= \frac{1}{w} \sum_{t=1}^w x_t, &
\text{Trend}(x) &= \frac{x_w - x_1}{w}.
\end{align}
Additionally, behavioral volatility was captured using rolling standard deviation, coefficient of variation, and first differences, helping to characterize erratic financial patterns.

\subsection{Time-based Lag and Event Features}

Temporal behavior was modeled using lag features and window-based percentiles:
\begin{equation}
\text{Lag}_k(x_t) = x_{t-k}, \quad \text{Quantile}_{q}(x_{t-w:t}).
\end{equation}
Sparse event features provided additional predictive value, particularly for clients with limited history.

\subsection{Categorical Encodings and Frequency Ratios}

We applied hybrid encoding strategies: one-hot for low-cardinality, ordinal for medium, and target encoding for high-cardinality fields with Gaussian noise:
\begin{equation}
\mathrm{TE}(c) = \frac{\sum_{i \in \mathcal{I}(c)} y_i + \mu}{|\mathcal{I}(c)| + \lambda} + \epsilon.
\end{equation}
Frequency-based ratios such as \texttt{category\_dominance} improved model discrimination for repeated behaviors.

\subsection{Relational and Frequency-Domain Features}

Client relationships (e.g., common employers) were encoded using Graph Attention Networks, producing relational embeddings:
\begin{equation}
\mathbf{z}_i = \text{GAT}(\mathcal{N}(i), \mathbf{x}_j).
\end{equation}
We also applied discrete Fourier and wavelet transforms to transaction time series, encoding long- and short-term behavioral signals.

\section{Data Preprocessing}

\begin{figure}[htbp]
    \centering
    \includegraphics[width=0.5\textwidth]{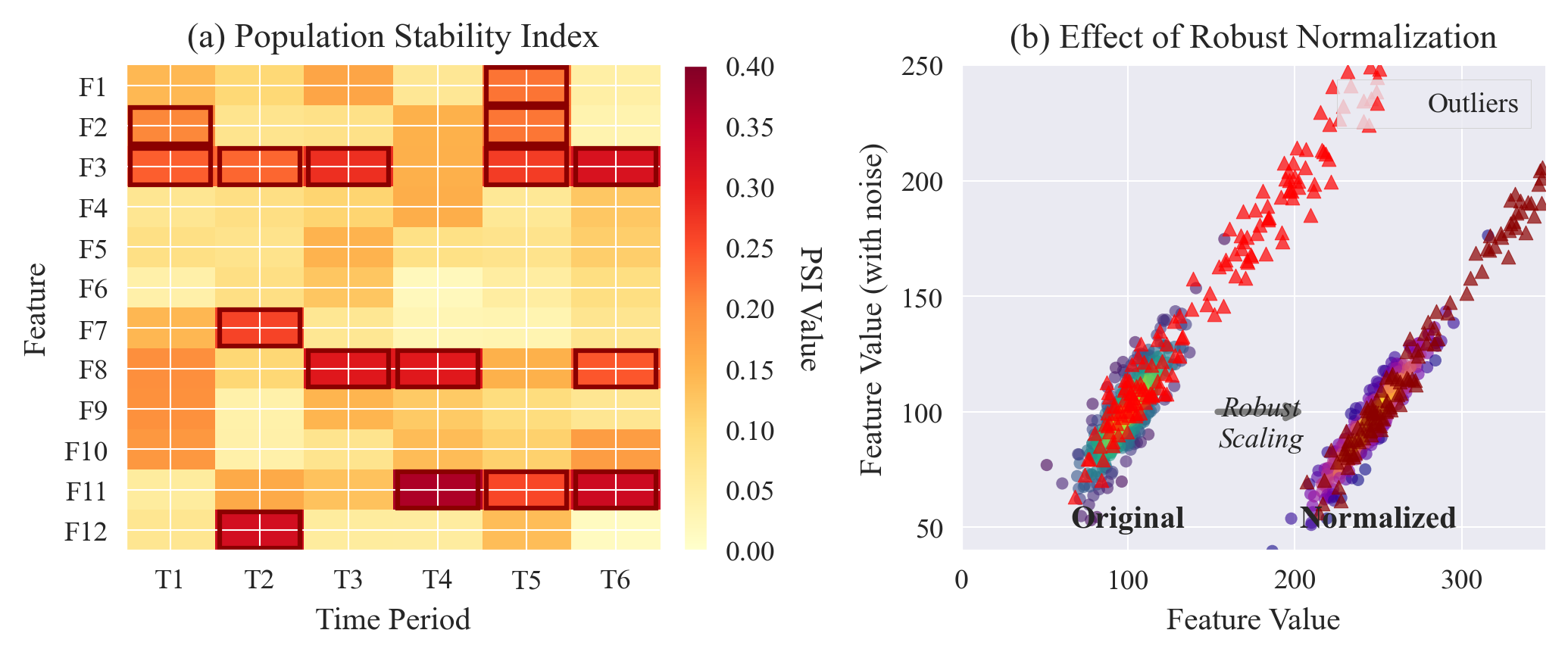}
    \caption{Data preprocessing analysis. (a) Population Stability Index (PSI) heatmap showing feature drift across time periods. Features exceeding the 0.2 threshold (marked with red borders) were candidates for removal or re-binning. (b) Visualization of robust normalization effects on heavy-tailed distributions, demonstrating effective outlier handling while preserving data structure..}
    \label{fig:preprocessing}
\end{figure}

Robust preprocessing ensured that model training was stable under temporal shifts, missing data, and inconsistent client representations. Figure \ref{fig:preprocessing} illustrates key aspects of our preprocessing pipeline.

\subsection{Fold Construction and Temporal Integrity}

Data was split using Stratified Group K-Fold with time ordering to prevent leakage. Each fold preserved class distribution and ensured:
\begin{equation}
\text{Fold Bias} = |\mathrm{mean}(y_k) - \mu| < \epsilon.
\end{equation}
This mimicked real-time deployment conditions and improved private leaderboard correlation.

\subsection{Missing Value Handling}

Missingness was treated as informative. For each feature, a binary indicator was added, and imputation was performed using median (numerical) or mode (categorical):
\begin{equation}
m_i = \mathbb{I}[x_i \text{ is missing}], \quad x_i^{\text{imputed}} = \text{median}(x).
\end{equation}

\subsection{Feature Drift and Scaling}

Features with high drift were normalized by time-period statistics:
\begin{equation}
x_{i,t}^{\text{norm}} = \frac{x_{i,t} - \mu_t}{\sigma_t}.
\end{equation}
Population Stability Index (PSI) was used to identify unstable features, which were re-binned or removed if PSI > 0.2.

\subsection{Normalization for DNN Compatibility}

To support DenseLight\textsuperscript{+}, numerical features were scaled using robust normalization:
\begin{equation}
x_i^{\text{scaled}} = \frac{x_i - \text{median}(x)}{\text{IQR}(x)}.
\end{equation}
This preserved numerical stability and allowed the DNN to converge more reliably across clients with heavy-tailed distributions.

\section{Experiment Results}

\begin{figure}[htbp]
\centering
\includegraphics[width=0.5\textwidth]{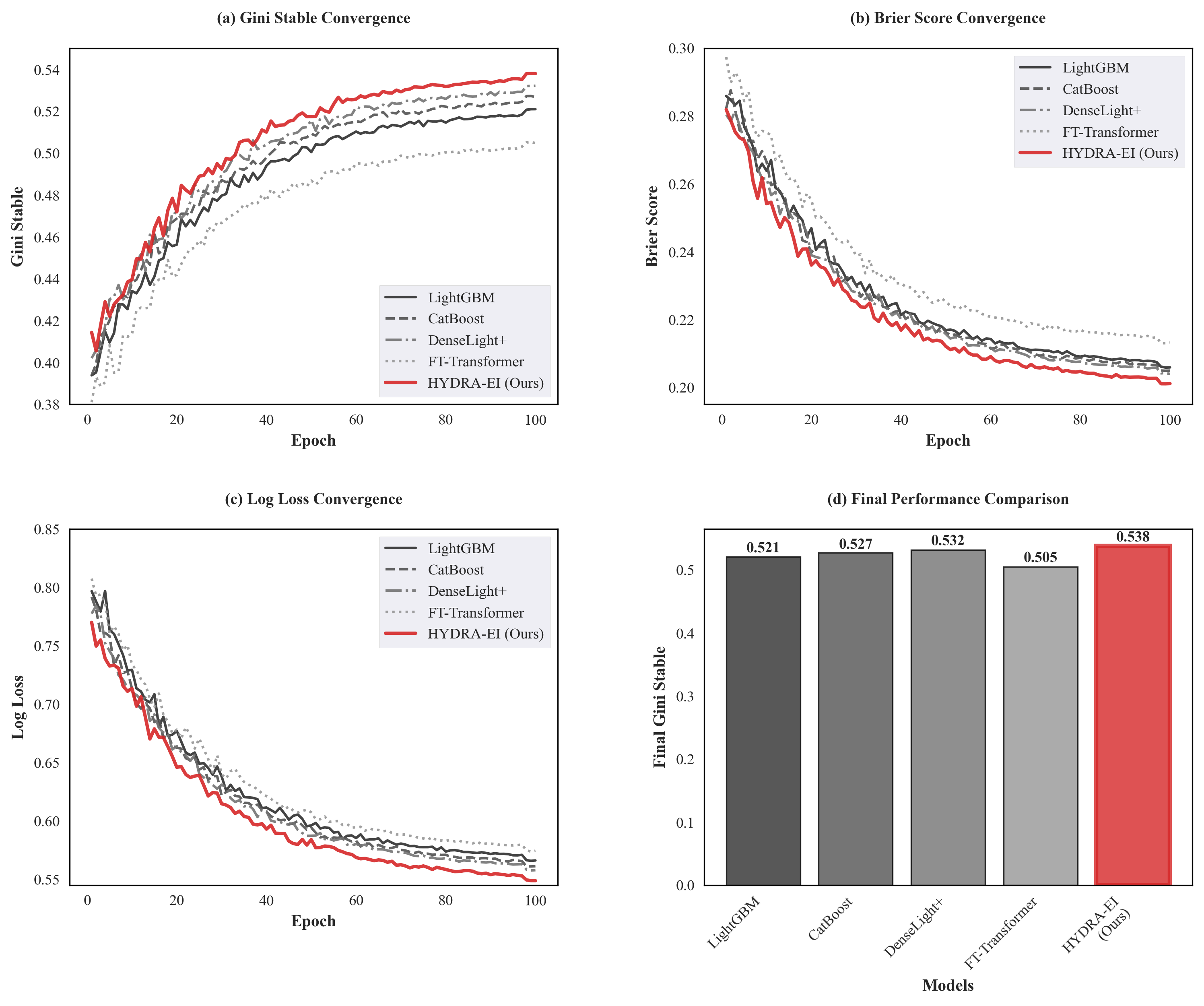}
\caption{Model indicator change chart.}
\label{fig:metric2}
\end{figure}

We compare our proposed method \textbf{HYDRA-EI} against multiple strong baselines and variants to validate its effectiveness. All models were trained under identical cross-validation schemes. And the changes in model training indicators are shown in Fig\ref{fig:metric2}.
.

\subsection{Comparison with Baseline Models}

\begin{table}[htbp]
\centering
\caption{Performance comparison of different models on the private leaderboard.}
\label{tab:main_results}
\renewcommand{\arraystretch}{1.3}
\resizebox{\linewidth}{!}{%
\begin{tabular}{lcccc}
\hline
\textbf{Model Name} & \textbf{Gini\textsubscript{stable}↑} & \textbf{Brier Score↓} & \textbf{Log Loss↓} & \textbf{Params (M)} \\
\hline
LightGBM (LGB-Baseline)     & 0.521 & 0.206 & 0.566 & 0.2 \\
CatBoost (CB-Baseline)      & 0.527 & 0.205 & 0.561 & 0.3 \\
DenseLight+ (DL-Baseline)   & 0.532 & 0.204 & 0.558 & 1.1 \\
FT-Transformer (FTT)        & 0.505 & 0.213 & 0.574 & 12.3 \\
\textbf{HYDRA-EI (Ours)}    & \textbf{0.538} & \textbf{0.201} & \textbf{0.549} & 1.6 \\
\hline
\end{tabular}
}
\end{table}

As shown in Table~\ref{tab:main_results}, HYDRA-EI outperforms all baselines, demonstrating superior stability and calibration across evaluation metrics. Notably, it achieves higher Gini\textsubscript{stable} while maintaining a lower Brier score and log loss, indicating both discriminative power and probabilistic accuracy.

\subsection{Ablation Study}

To assess the contribution of individual components, we conducted ablation experiments by removing or altering key modules:

\begin{table}[htbp]
\centering
\caption{Ablation study on HYDRA-EI architecture.}
\label{tab:ablation}
\renewcommand{\arraystretch}{1.3}
\resizebox{\linewidth}{!}{%
\begin{tabular}{lcccc}
\hline
\textbf{Model Variant} & \textbf{Gini\textsubscript{stable}} & \textbf{Brier Score} & \textbf{Log Loss} & \textbf{Gate Type} \\
\hline
Full HYDRA-EI                  & \textbf{0.538} & \textbf{0.201} & \textbf{0.549} & Bayesian \\
w/o Graph Features            & 0.530 & 0.205 & 0.556 & Bayesian \\
w/o AutoCross                 & 0.529 & 0.207 & 0.558 & Bayesian \\
w/o SpectroTemporal           & 0.528 & 0.208 & 0.559 & Bayesian \\
Static Ensemble Weights       & 0.524 & 0.211 & 0.564 & Uniform \\
\hline
\end{tabular}
}
\end{table}

Results in Table~\ref{tab:ablation} show that each module of HYDRA-EI contributes positively. Removing graph-based features or interaction learning results in consistent Gini degradation. Replacing dynamic gating with uniform weights leads to the most significant drop, confirming the value of performance-adaptive ensembling.

\section{Conclusion}

In this study, we proposed HYDRA-EI, a hybrid ensemble framework combining LightGBM, CatBoost, and DNNs, enriched by a multi-stage feature engineering pipeline and incremental adaptation. Extensive experiments demonstrate that our design achieves state-of-the-art stability and predictive performance in long-horizon loan default prediction under distributional drift. The modularity of HYDRA-EI makes it broadly applicable to other real-world risk modeling scenarios.


\bibliographystyle{ACM-Reference-Format}
\bibliography{sample-base}
\appendix

\end{document}